# Embedding Java Classes with code2vec: Improvements from Variable Obfuscation

Rhys Compton
Eibe Frank
rhys.compton@gmail.com
eibe.frank@waikato.ac.nz
University of Waikato
Hamilton, New Zealand

Panos Patros
Abigail Koay
panos.patros@waikato.ac.nz
abigail.koay@waikato.ac.nz
University of Waikato
Hamilton, New Zealand

## ABSTRACT

Automatic source code analysis in key areas of software engineering, such as code security, can benefit from Machine Learning (ML). However, many standard ML approaches require a numeric representation of data and cannot be applied directly to source code. Thus, to enable ML, we need to embed source code into numeric feature vectors while maintaining the semantics of the code as much as possible. **code2vec** is a recently released embedding approach that uses the proxy task of method name prediction to map Java methods to feature vectors. However, experimentation with **code2vec** shows that it learns to rely on variable names for prediction, causing it to be easily fooled by typos or adversarial attacks. Moreover, it is only able to embed individual Java methods and cannot embed an entire collection of methods such as those present in a typical Java class, making it difficult to perform predictions at the class level (e.g., for the identification of malicious Java classes). Both shortcomings are addressed in the research presented in this paper. We investigate the effect of obfuscating variable names during training of a **code2vec** model to force it to rely on the structure of the code rather than specific names and consider a simple approach to creating class-level embeddings by aggregating sets of method embeddings. Our results, obtained on a challenging new collection of source-code classification problems, indicate that obfuscating variable names produces an embedding model that is both impervious to variable naming *and* more accurately reflects code semantics. The datasets, models, and code are shared[1] for further ML research on source code.

---

[1]https://github.com/basedrhys/obfuscated-code2vec

## 1 INTRODUCTION

Analysing source code using machine learning has a range of uses. In software engineering it can improve the readability of decompiled code by suggesting method names that have been lost, and improve source code search; in authorship identification, it can help detect whether two pieces of code were written by the same person; and in malware detection, it can be used to identify potentially malicious code.

Typical Machine Learning (ML) approaches require the input data to be in a numerical form. Natural Language Processing (NLP) has best addressed this problem with word embeddings: numerical vector representations of words such that the vector representing a word preserves its meaning, with semantically similar words being mapped to similar vectors [18]. **code2vec** [5], which this research is based on, similarly uses a novel neural network-based approach for learning code embeddings by employing the proxy task of method name prediction, for which labelled training data is plentiful. **code2vec** converts Java code into a set of paths from the code's Abstract Syntax Tree (AST) and learns to combine these paths using an attention mechanism.

In NLP, large-scale pretraining of word embeddings, which are then used as input for more complex ML models in downstream tasks [18], enables these models to leverage the semantic information held in the embeddings. Using code embeddings has a similar benefit: embeddings can be trained that capture the semantics of source code, then used for a range of down-stream tasks (e.g., code author identification and malware classification) with minimal fine-tuning and far less training data than would be necessary to train a model for these tasks from scratch.

The work we present is motivated by an empirically observed weakness in the original application of **code2vec**: the AST contains variable names, which the **code2vec** model leverages heavily for prediction. This heavy dependency poses a problem for the model's generalisation: different developers may name their variables differently, developers from different countries likely name variables in their native language, and malware creators often adversarially obfuscate their code to avoid detection. In these cases, the model may be mislead and mispredict the code's functionality. Indeed, our experiments show that this over-reliance on variable names can negatively affect classification accuracy when applying the embedding model to tackle algorithm identification and similar source code classification tasks. To address the shortcomings of the methods and techniques implemented in **code2vec**, we empirically examine the effectiveness of two code obfuscation approaches that attempt to force the embedding model to learn from the code structure rather than variable names. Another drawback of this state-of-the-art method is that it can only create embeddings for individual functions and cannot summarise an entire Java class in a single vector. We evaluate the efficacy of simple mathematical operations on sets of vectors to address this shortcoming. Our experiments show which aggregation methods provide the most descriptive summary vector for a prediction task.

For our investigation, we employ a broad range of source code classification tasks and, using the **code2vec** models as feature extractors, compare classification accuracy obtained with embeddings from the obfuscated and non-obfuscated **code2vec** models; the observed accuracy indicates how informative the embeddings are for the corresponding domain.

There has been recent complementary work evaluating code2vec embeddings [16], specifically the generalizability of the *token embeddings*. Their work found that token embeddings in isolation were not suitable for the downstream tasks tested [16], which is reasonable when considering that the attention mechanism present in **code2vec**—a significant part of the model—is not used. Their findings also agree with our results in that, when discarding token embeddings that represent variable names, the resulting representation is often superior.

The research questions we attempt to answer are:

- *Does obfuscation of variable names yield an improved model of code semantics?*
- *How should we aggregate embeddings for methods in a Java class to accurately describe the class as a single entity?*

The experimental results we present show that randomly obfuscating variable names yields a more robust **code2vec** model that is less reliant on variable names than the original **code2vec** model. Our results also show that aggregating *all* methods in a class using *mean*, *meanMin* or *meanStd* retains the most information about the class. In our experiments, we also consider dimensionality reduction using the UMAP technique [17] but find that this harms classification accuracy.

The main contributions of this research are:

- An investigation of obfuscating variable names during training and testing of a **code2vec** embedding model.
- An evaluation of **code2vec** embeddings on a wide range of tasks requiring classification of source code.
- Baselines for creating embeddings for files containing many methods, by applying simple aggregation functions on the individual method embeddings.
- Collation of a wide range of Java source code classification datasets.

## 2 LEARNING CODE EMBEDDINGS

The **code2vec** model is based on a method of extracting information from an Abstract Syntax Tree (AST) constructed from the code.

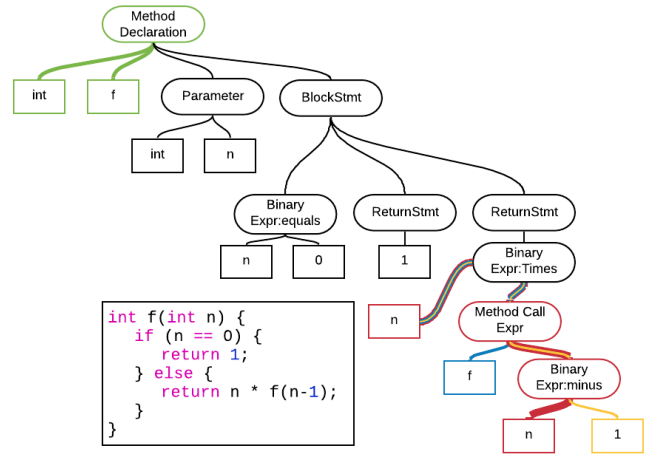

**Figure 1: Example code for a function calculating the factorial and the parsed AST.**

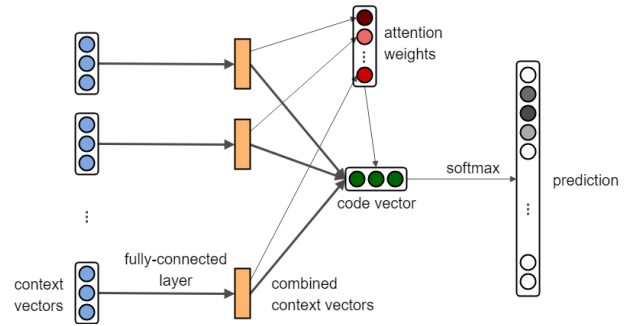

**Figure 2: code2vec model architecture [5].**

An AST is a tree-like representation of a piece of source code. It does not contain every detail in the original source code (e.g., parentheses and comments) but rather, represents the important syntactic structures that make up the functionality of the code.

Figure 1 shows an example of an AST for a recursive `factorial` function. This tree structure is a more abstract alternative to representing the code as a flat sequence of characters or tokens, an approach taken in early work on applying ML to code [1], [2], [22]—more recent approaches have used ASTs to represent code, showing that this method leverages the highly structured syntax of source code, improving model training time and accuracy [4].

**code2vec** [5] is a neural attention-based approach to learn embeddings for code snippets (e.g., individual methods in a Java class). The aim is to obtain a *code embedding* such that the semantics of the code are preserved as much as possible in the embedding (i.e., semantically similar methods are mapped to similar vectors). Based on the strong results in [4], **code2vec** represents code snippets as a set of *path-contexts*: paths between nodes on the AST constructed from the code snippet.

Figure 2 shows the architecture of the neural network used in **code2vec**. It is trained by optimising classification accuracy on the proxy task of predicting method names based on the code of the methods. The method is explained succinctly below; for more information refer to the original paper [5]. To explain the method more succinctly, it is helpful to use the terminology introduced in the original **code2vec** paper [5]:

***AST Path:*** Given an AST, an *AST Path* is a path between two leaf nodes in this AST. The **red**, **green**, **yellow**, and **blue** coloured paths in Figure 1, highlight some example AST paths.

***Path-Context:*** Given an AST Path $p$, its path-context is the triplet $\langle x_s, p, x_t \rangle$, with $x_s$ and $x_t$ being the start and end nodes of $p$ respectively. In other words, the path-context describes two tokens corresponding to two AST leaf nodes and the path between them.

*Example* [5]: A potential path-context that represents the statement: x = 7;
$\langle x, (NameExpr \uparrow AssignExpr \downarrow IntegerLiteralExpr), 7 \rangle$

A $\uparrow$ delineates moving towards the root of the AST; whereas a $\downarrow$ moving towards the leaves.

***AST Parsing:*** The analysis of code in **code2vec** has two main steps: AST parsing and path-context selection. The code is parsed into an AST (Figure 1) and then converted into a set of path-contexts. Path-contexts are created from every leaf to every other leaf.

***Path-Context Selection:*** The start and end nodes, and the path between them are each represented by a vector whose values are learned through training and concatenated into a *path-context vector* (PCV). During training, the *attention mechanism* in the model learns how to combine these PCVs, putting more focus on certain PCVs (i.e., parts of the code). This enables the model to identify parts of the AST that are highly predictive (e.g., certain variable names) and ignore parts that are not. Empirically, it has been shown to be able to discern subtle differences between similar Java methods [5].

Unfortunately, the strength of the attention mechanism means the model can be fooled easily. The authors of **code2vec** note the model's dependency on variable names and the loss of accuracy on method name prediction when given obfuscated or adversarial variable names. This detrimental effect is due to the model being trained on top-starred GitHub projects where variable names follow best practice and so are highly predictive; thus, the model learns to leverage them [5]. To illustrate this problem, Figure 3 shows two methods with changed variable names (due to a typo in the case of the first method, and a reasonable alternative name in the second method) and the difference in predicted method name.

To illustrate how misleading the model predictions are in these cases, we can directly compare the embeddings obtained for each of the two versions of the two methods by measuring cosine similarity and Euclidean distance between the vectors. Cosine similarity is a measure of how similar the angles of two vectors are. It ranges from -1 (exactly opposite), 0 (orthogonal), to 1 (parallel). Using this, the embeddings for the `factorial` example in Fig. 3 have a similarity of only **0.644** and Euclidean distance of 7.4, while the done embeddings are even more dissimilar with a cosine similarity of **0.239** and distance of 12.7. Altering a variable name has a detrimental effect on code2vec's representation accuracy; in both cases, altering a name yields misleading embeddings and entirely incorrect output predictions of even higher confidence than with correct names.

```
void f() {
    boolean done = false;
    while (!done) {
        if (remaining() <= 0) {
            done = true;
        }
    }
}
```
(Correct) Predictions:
**done**        34.27%
**isDone**      29.79%
**goToNext**    12.81%
**current**     8.93%

```
void f() {
    boolean don = false;
    while (!don) {
        if (remaining() <= 0) {
            don = true;
        }
    }
}
```
(Incorrect) Predictions:
**createMessage**  75.07%
**checkMessage**   16.25%
**compareTo**      8.55%
**putMessage**     0.06%

```
int f(int n) {
    if (n == 0) {
        return 1;
    } else {
        return n * f(n-1);
    }
}
```
(Correct) Predictions:
**factorial**  47.73%
**fact**       22.99%
**fac**        9.15%
**spaces**     7.11%

```
int f(int total) {
    if (total == 0) {
        return 1;
    } else {
        return total * f(total-1);
    }
}
```
(Incorrect) Predictions:
**getTotal**  84.21%
**total**     4.06%
**average**   2.31%
**setTotal**  2.25%

**Figure 3: Changes in variable names producing significantly different predictions. A typo of `done` (top), and an alternative name, changing `n` to `total` (bottom).**

## 3 OBFUSCATION OF VARIABLE NAMES

We now outline the two variable obfuscation methods we designed for our experiments with the aim of mitigating the problems discussed in Section 2. In particular, the methods are applied to the training data of the **code2vec** model under the hypothesis that hiding informative variable names will force the model to learn from the structure of the code rather than particular variable names. Leveraging variable names for method name prediction introduces a strong bias and can cause the model to generalise poorly.

The two obfuscation methods we propose are:

- **Type Obfuscation** - Variable names are replaced with a string indicating the scope and type of the variable.
- **Random Obfuscation** - Variable names are replaced with a random string of letters.

Figure 4 shows an example Java method and the result of applying each type of obfuscation.

*Type obfuscation* was used as some information about the variable is retained in the name; the model can still see what type the variable is and what scope it is—both pieces of information that are potentially useful for prediction. Types can usually be ascertained from Java source code analysis, however, `unk` was used as the type in rare cases where this failed.

*Random obfuscation* was used as a method to completely remove reliance on variable names by replacing them with a randomly generated string of letters. This was done so the model will not be able to learn any trends from variable names (as they are all

```java
// Original method
public String getResult(String input) {
    int count = this.objCount;
    this.objCount++;
    return input + Integer.toString(count);
}

// Type Obfuscated method
public String getResult(String param_string_1) {
    int local_int_1 = this.field_int_1;
    this.field_int_1++;
    return param_string_1 + Integer.toString(local_int_1);
}

// Random Obfuscated method
public String getResult(String FDGHSXJF) {
    int UJGRJWMU = this.MVKHEQTR;
    this.MVKHEQTR++;
    return FDGHSXJF + Integer.toString(UJGRJWMU);
}
```

Figure 4: Obfuscated versions of a Java method.

random), and consequently will rely more on the underlying code structure than when applying type-obfuscation.

When considering obfuscation methods, it is worth inspecting their effect on the path-contexts as these are what is ultimately used in the model. In a path-context, the **code2vec** model learns an embedding for the start leaf-node, the path, and the end leaf-node and concatenates them into a single *path-context vector*, e.g., for the path-context shown earlier in Section 2, embeddings are learned for x, (*NameExpr* ↑ *AssignExpr* ↓ *IntegerLiteralExpr*), and 7, and concatenated into a single vector. Due to the structure of ASTs, variable names always appear in the leaf nodes; hence, one might ask why not simply remove the leaf-node embeddings and use the path embedding to solely represent the entire path-context. This harms accuracy in the task of method name prediction: the **code2vec** authors established this by performing an ablation study in which they compared prediction accuracy after removing different parts of the path-context vector. A key observation is that although all variable names appear in leaf nodes, not all leaf nodes are variable names; for the path-context above, removing leaf-node embeddings *would* remove reliance on the variable name x but also hide the embedding for 7, which may hold useful information about the code's semantics.

The random obfuscation method solves this problem: randomly obfuscating variable names should remove the dependency on leaf-node embeddings when they are variable names, as the random out-of-vocabulary tokens are replaced with UNK by the model; however, it still allows the model to learn from leaf-node embeddings when they represent other constructs (return values, method names, assigned values, etc.) and the path between the nodes still holds information. Therefore, random obfuscation is distinctly different from what was evaluated in the ablation study.

In theory, type-obfuscation provides a middle ground: variable name embeddings can be informative (e.g., param_string_1) but the original variable names (e.g., inputWord), which can vary depending on the code's author, are hidden.

It is also important to note that the **code2vec** authors' ablation study only examined the effect of removing information at *test* time, i.e., the model was trained with the full path-context (full information) then tested with parts of the path-context removed (partial information), understandably getting poor results as the information the model learned to leverage was no longer available. Conversely, this research examines the effect of hiding information at *training* and test time, giving the model a chance to learn around the hidden variable names and leverage the information that is left.

Implementation of our obfuscation methods is straightforward; a command-line tool was created with Spoon[2] that walks through a directory containing Java files and replaces variable names in a copy of the files using the specified obfuscation method. The .jar tool and source code are free for use with the rest of the paper code.

## 4 TRAINING AND EVALUATING CODE2VEC EMBEDDINGS

This section outlines how each **code2vec** model was trained and how the embeddings were evaluated. A range of datasets were collected for this research, which are also explained.

### 4.1 Training code2vec Models

The dataset used for training the **code2vec** models in our experiments was java-large, sourced from the **code2seq** [3] GitHub repository[3], a project related to **code2vec** that considers translating source code to a descriptive sequence of words and provides a larger dataset (~15.3m examples) than the original **code2vec** dataset (~14m examples).

Four embedding models were used in our comparison:

**Standard baseline:** A **code2vec** model trained on the full java-large dataset with no obfuscation. (*~15.3m training examples; No obfuscation.*)

**Type obfuscated:** The variable names in the dataset for this model are hidden using the type obfuscation method. The obfuscation tool could not process some files due to peculiarities in the files' contents, which reduced the size of the obfuscated datasets. (*~13.7m examples; Type obfuscation.*)

**Random obfuscated:** The variable names in the dataset for this model are hidden using the random obfuscation method. The same reduced training set size applies in this case as well. (*~13.7m examples; Random obfuscation.*)

**Reduced:** Due to the smaller size of the obfuscated datasets, the Reduced model is created to control for this by being trained on a non-obfuscated dataset of the same size as the obfuscated datasets. (*~13.7m examples; No obfuscation.*)

Each of the four models were trained using the training script supplied in the GitHub repository for **code2vec** with default hyperparameters, and trained until the validation set performance stopped increasing (3 epochs in our experiment).

### 4.2 Dataset Pipeline

In addition to the four different embeddings models, we consider several aggregation strategies to form class-level embeddings by combining the method-level embeddings from a Java class into a single vector. This creates a large configuration space for the experiments. To enable systematic exploration, we created a *dataset pipeline* as a way to compare the different embedding models and configurations for aggregating vectors. Figure 5 shows the high level idea: given a labelled dataset of Java source code files, use a trained **code2vec** model as a feature extractor, creating embeddings for the code and converting it into a numerical format that can be used in any standard machine learning framework (WEKA [13] in this case). The quality of the embeddings can then be measured indirectly by considering the estimated accuracy on the resulting classification problem (e.g., how accurately the WEKA classifier can differentiate between the two types of Java class).

---
[2]https://github.com/INRIA/spoon
[3]https://github.com/tech-srl/code2seq

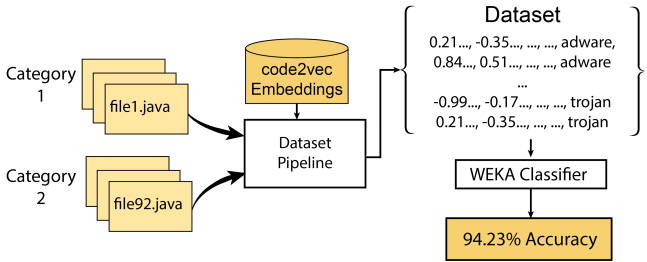

Figure 5: High-level view of pipeline.

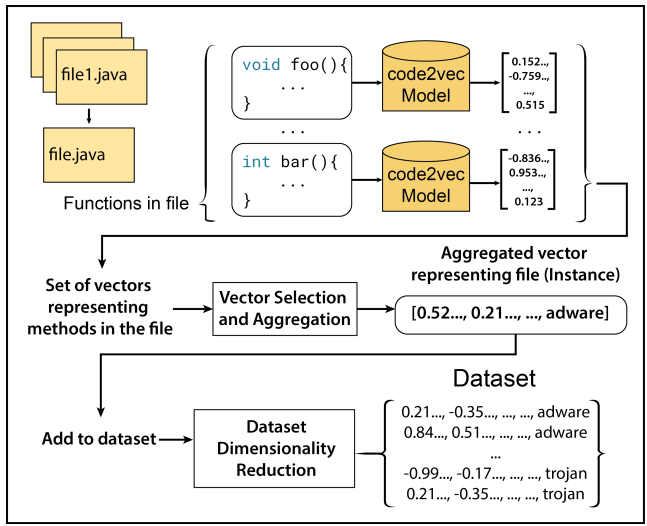

Figure 6: Detailed dataset pipeline.

In addition to the models that are compared, there are three pipeline parameters that can be changed: the selection method, the aggregation method, and whether dimensionality reduction is performed. The place of these within the pipeline can be seen in Figure 6. They are discussed in Sections 4.4, 4.5, and 4.6 respectively.

### 4.3 Extracting Embeddings

First, each file (i.e., Java class) is deconstructed into its individual methods and each method converted into a code embedding using the supplied **code2vec** model. The set of embeddings is then passed to the next step of the pipeline to be aggregated.

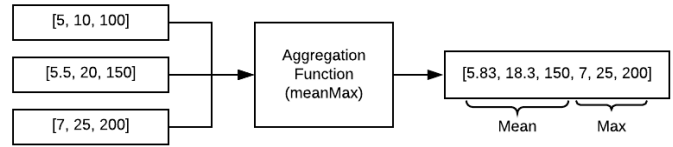

Figure 7: Example aggregation function of *meanMax* on three small vectors. Note that this can scale to any number of vectors of any size, as long as each vector is of the same length.

### 4.4 Selection Method

It is possible that a few of the vectors representing methods in a file are not useful when describing the semantics of the file as a whole. This gave rise to the idea of experimenting with different selection methods prior to aggregation of the embeddings. This idea is loosely inspired by the concept of feature selection: irrelevant features in a dataset can negatively impact model performance, therefore removing them may improve the model [9], [21].

Three selection methods were evaluated: **All** - Select all methods in a file; **Top K** - Select the top $K$ methods in a file, ordered by length (number of lines in the method), under the hypothesis that longer methods contain more information about the class than shorter ones; **Random K** - Randomly select $K$ methods from the file.

### 4.5 Aggregation Method

After selection, the aggregation method receives the selected set of vectors. The task now is to aggregate them into a single vector that represents the entire file. This is similar to the task in **code2vec**, in which the model needs to aggregate a set of PCVs into a single vector representing the code snippet. The attention-based method for achieving this in **code2vec** was not investigated for the task of aggregating method embeddings due to the lack of appropriate datasets to train the attention mechanism; in contrast, the simple aggregation methods we experimented with do not require any training and can be applied to any number of examples.

The aggregation methods are applied *column-wise*, taking in a set of numbers from a column, applying the function, and then returning the result as the value for this column in the resulting vector (e.g., Figure 7). The base aggregation functions used were *max*, *min*, *sum*, *mean*, *median*, and *standard deviation*. The powerset up to size 2 was used for the list of all aggregation functions tested, i.e., in addition to each individual aggregation function, all combinations of size two were tested (Max-Min, Max-Sum, Max-Mean, etc). For a combination of aggregation functions, the two functions were simply concatenated together. Preliminary testing showed *min, mean,* and *max* to be strong functions (good prediction accuracy) so the extra function *minMeanMax* was also tested, as well as a combination of all aggregation methods (*minMaxMeanMedianStddevSum*). This gave a list of 23 different aggregation functions.

### 4.6 Dimensionality Reduction

Dimensionality reduction was tested to see if reducing the number of features in the resulting dataset can improve prediction accuracy, with UMAP [17] ultimately being used as the method of reduction.

Although UMAP is commonly used for visualisations, the authors note it is a "...general purpose dimension reduction technique for machine learning..."[17], and it has been used for general reduction previously [8]. PCA was considered for this task but is a linear technique and has negligible effect when used with a linear model such as what was used in our experiments.

## 5 COLLATING EVALUATION DATASETS

To test the predictive quality of the code embeddings created by each **code2vec** model, a number of datasets were gathered. These aim to both test how accurate these embeddings can be for predictive tasks and act as different benchmarks to compare the obfuscated and non-obfuscated models.

There are seven datasets collated for the evaluation:

(1) OpenCV/Spring - 2 categories, 305 instances
(2) Algorithm Classification - 7 categories, 182 instances[4]
(3) Code Author Attribution - 13 categories, 1062 instances
(4) Bug Detection - 2 categories, 31135 instances[5]
(5) Duplicate File Detection - 2 categories, 1669 instances[6]
(6) Duplicate Function Detection - 2 categories, 1277 instances
(7) Malware Classification - 3 categories, 20927 instances[7]

In experimentation, the datasets were limited to 2000 instances per class by random sampling to speed up embedding creation and reduce class imbalance in the larger datasets. Datasets (1) and (3) are newly created for this research. (2), (4), and (7) use existing data transformed into multi-class classification datasets. These datasets have not been used before in the ML literature in this transformed representation.

### 5.1 OpenCV/Spring

As an initial task, we created a dataset to differentiate between two types of Java files: those using OpenCV, an open-source image processing library, and those using Spring, a popular Java web application framework. These two classes of files were chosen as they represent two contrasting tasks (image processing vs. serving web pages) and so, a poor result on this dataset would mean the model struggles even on tasks where obtaining discriminative embeddings should be quite straightforward. This dataset was created manually using files from GitHub and other website tutorials.

### 5.2 Algorithm Classification

Princeton University has made public at their *Java Algorithms and Clients* page a number of algorithm implementations in Java, conveniently sorted into their respective categories (fundamentals, sorting, searching, etc.), which were used for this dataset.

### 5.3 Code Author Attribution

An area of research in code analysis is *Code Author Attribution*, the task of identifying the most likely author of a piece of code given a set of predefined authors [14] [10]. Advances in this area have strong implications for cyber-security, privacy, and plagiarism detection; effective methods could be used to tell if a student has plagiarised their assignment, whether a developer has violated a non-compete clause, or even who has created a specific type of malware.

Leetcode[8] is a website designed for people to improve their coding skills by solving questions posted on the site. People commonly publish their Leetcode solutions on GitHub as a proof of talent and to help others, so solutions for these common tasks were sourced from GitHub to create a code authorship dataset, as there are different authors solving the same set of problems.

We consider this an *anti-task*, a task requiring a polar opposite approach to the goal targeted by training the embedding models: for two pieces of code written by different people, if they are semantically similar, the model was trained to embed them into the same point in embedding space. In other words, it was trained to ignore subtleties between authors and embed their code into the same place. Because of this model property, distinguishing between authors based on **code2vec** embeddings *should* be a difficult task.

### 5.4 Bug Detection

Bug detection is the task of detecting bugs in code, based on hand-crafted features or from the source code itself. The Public Unified Bug Dataset for Java [12] is a dataset of ~43k Java files compiled from five separate datasets, hand-labelled with the number of bugs that occur in the file. For the embedding evaluation, this dataset was altered to be a 2-class classification problem: files labelled as having *no bugs* or *at least one bug*.

### 5.5 Duplicate Detection

Duplicate detection is an important task as duplicate code is usually a sign of poor implementation and can harm the maintainability of software systems [6]. Source{d}, a software company, has shared their hand-labelled duplicate dataset on GitHub, which contains pairs of both methods and files classified as either *yes*, *maybe*, or *no* to being duplicates. Labelling duplicate code is intrinsically a subjective task and having a *maybe* category makes the labels in the dataset even more subjective, so this category was removed. Rather than creating and aggregating embeddings for a single class and using that for the resulting dataset, the embeddings for both pieces of code in the duplicate pair were created and aggregated, and then subtracted from each other. This means a resulting instance for a pair of code that is exactly the same, has attribute values ~0: both items in the pair create the same embeddings, so subtracting them from each other results in 0 for each attribute. As the dataset contained pairs of both methods and files, separate duplicate method and duplicate file datasets were created.

### 5.6 Malware Classification

Malware classification research [19], [15], in the context considered here, attempts to develop methods to automatically classify a piece of malicious code into one of a set of types of malware (trojan, adware, etc). The Android Malware Dataset [20] is a collection of ~24k malware samples that belong to 10 malware types. The instances come as .apk files so were decompiled into Java using JADX[9]. Note that the decompilation process was able to recover most of the original source code, however, for some files variable names were not recovered. From the 10 malware types, three were arbitrarily chosen for this dataset: adware, ransomware, and trojan.

---

[4] https://algs4.cs.princeton.edu/code/
[5] http://www.inf.u-szeged.hu/~ferenc/papers/UnifiedBugDataSet/
[6] https://github.com/src-d/datasets/tree/master/Duplicates
[7] http://amd.arguslab.org/
[8] https://leetcode.com/
[9] https://github.com/skylot/jadx

**Table 1: code2vec F1 Score on validation set and test set for method name prediction.**

|         | Std   | Reduc | Type-obf | Rand-obf |
|---------|-------|-------|----------|----------|
| Val Set | 0.399 | 0.396 | 0.337    | 0.355    |
| Test Set| **0.433** | 0.431 | 0.389    | 0.386    |

## 6 EXPERIMENTAL RESULTS

We first consider method name prediction performance by each **code2vec** model—i.e., classification performance in the task used for training each embedding model—and then evaluate which pipeline configuration results in the most informative file embeddings, how informative the embeddings are in the different domains, and the effect of obfuscating test data for a non-obfuscated **code2vec** model.

### 6.1 Method Name Prediction

We first consider performance of the four neural networks discussed in Section 4.1 that yield the four different **code2vec** embeddings we consider in our experiments. Here, performance is measured using the F1 score as reported by the **code2vec** model: the harmonic mean of precision and recall, on the method name prediction task used to train the embedding model. Table 1 shows the maximum F1 score achieved by each of our four **code2vec** models during training on their respective datasets' validation set, and the F1 score of the chosen model on the corresponding separate test set.

*Standard* achieved the best F1 score on both the validation set and test set with scores of 0.399 and 0.433 respectively, while the *Reduced* model was minimally worse but effectively the same. The two obfuscated models saw a clear decrease in performance, achieving an F1 score 0.04 (~10%) less than the full non-obfuscated model.

These results show that both forms of obfuscation reduce performance when predicting method names, which matches the ablation study by the code2vec authors, and was an expected result when considering the naming information held in variables and the high frequency of get/set methods in the training dataset: the name of a simple get method (e.g. getName) can only be accurately predicted when the name of the variable it is returning is visible. We now move into evaluation of the *embeddings* created by the four models through the use of the *dataset pipeline*.

### 6.2 Selection Methods

The set of vector selection methods discussed in Section 4.4 were tested on the OpenCV/Spring and Algorithm datasets, using embeddings from the *Reduced* model. Preliminary tests showed a linear support vector machine classifier trained with WEKA's implementation of LibLinear [11] obtained the best classification performance on datasets created from these embeddings, so all following results are based on LibLinear. LibLinear was applied in default mode, generating a support vector machine based $L_2$-regularized $L_2$ loss; the regularisation parameter $C$ was left at its default value 1.

**Table 2: Comparing selection methods by average kappa over 10 runs of 10-fold cross-validation with LibLINEAR, over the 23 aggregation methods, using embeddings from the Reduced model. Highest kappa bolded.**

| Selection | OpenCV/Spring | Algorithm Classification |
|-----------|---------------|--------------------------|
| All       | 0.923         | 0.555                    |
| Rand 1    | 0.823         | 0.356                    |
| Rand 2    | 0.898         | 0.440                    |
| Rand 3    | 0.912         | 0.493                    |
| Rand 5    | 0.915         | 0.527                    |
| Top 1     | 0.918         | 0.412                    |
| Top 2     | **0.926**     | 0.536                    |
| Top 3     | 0.923         | 0.558                    |
| Top 5     | 0.925         | **0.563**                |

For the performance statistics, kappa ($\kappa$) is reported herein out. Kappa is an improvement on % accuracy as it shows improvement in classification accuracy relative to a random classifier [7]. This makes comparisons between datasets more informative than simple % accuracy, especially for datasets with a class imbalance or different numbers of classes; in terms of real-world performance, 50% accuracy on a 2-class dataset is not equivalent to 50% accuracy on a 10-class dataset, however, kappa will reflect this difference.

Table 2 shows classification accuracy of LibLinear measured by the $\kappa$ statistic, as estimated using 10 runs of stratified 10-fold cross-validation. Note that each selection method was evaluated in conjunction with each of the 23 aggregation methods discussed in Section 4.5 so each result is an average of 10×10×23 estimates.

Overall, the best performing selection methods were *All*, and *Top K* with $K = [2, 5]$, but the difference between these is marginal. The Algorithm Classification dataset exhibited a larger difference between best and worst selection method, with a difference in $\kappa$ of 0.207, while there was a difference of only 0.102 for OpenCV/Spring. This higher difference is likely due to the number of methods in each class in the Algorithm Classification dataset—files tended to have more methods, so more variation can be expected between selection methods; OpenCV/Spring instances had fewer methods per class and so not much difference was observed.

Comparing *Top K* to *Random K*, the former performed better overall. This supports the hypothesis that *longer methods contain more information*: the higher classification accuracy of *Top K* compared to *Random K* at lower values of $K$ (e.g., *Rand 1* vs *Top 1*) indicates that selecting the longest method from a Java class is more informative for describing the class/file than random selection. However, neither of these are substantially better than simply applying no selection (use all methods to describe the class); consequently, we use this method in all subsequent experiments.

### 6.3 Dimensionality Reduction

To evaluate the effect of dimensionality reduction on the embeddings, we used UMAP to reduce their size from 384 to 25, 50, 100, and 250 dimensions respectively. The results in Table 3 show the average $\kappa$ over all 23 aggregation methods, over 10 runs of stratified 10-fold cross-validation with LibLinear. For reasons mentioned in Section 6.2, *All* was used as the selection method.

Table 3: Comparing dimensionality reduction levels, with the average kappa over 10 runs of 10-fold cross-validation with LibLINEAR, over the 23 aggregation methods, using embeddings from the Reduced model.

| Reduction | OpenCV/Spring | Algorithm Classification |
|---|---|---|
| **None** | **0.924** | **0.672** |
| UMAP - 25 | 0.876 | 0.526 |
| UMAP - 50 | 0.869 | 0.537 |
| UMAP - 100 | 0.884 | 0.510 |
| UMAP - 250 | 0.882 | 0.529 |

Table 4: Example scoring for algorithm classification.

| | Obfuscation Type | | | | |
|---|---|---|---|---|---|
| Agg Function | None | Type | Random | Average | Score |
| maxMed | 0.699 | 0.751 | 0.758 | 0.736 | 5 |
| minMeanMax | 0.688 | 0.751 | 0.764 | 0.734 | 4 |
| medStd | 0.705 | 0.724 | 0.762 | 0.730 | 3 |
| maxMin | 0.673 | 0.764 | 0.750 | 0.729 | 2 |
| meanStd | 0.701 | 0.726 | 0.756 | 0.728 | 1 |

The results show that performing no dimensionality reduction gave the best prediction accuracy with *LibLINEAR*. There was a ~0.05 drop in $\kappa$ for the OpenCV/Spring dataset when UMAP was applied, while the Algorithm Classification dataset exhibited a more significant ~0.16 decrease. The extent of dimensionality reduction showed little effect on overall embedding quality.

Based on the results presented above, it was found unnecessary to perform method selection or dimensionality reduction, and so these hyperparameters of the *dataset pipeline* were no longer explored: only the baseline methods are used from now on.

### 6.4 Aggregation Methods

We now evaluate the individual 23 aggregation methods on all seven datasets. For each of the seven datasets, the 23 aggregation methods were sorted in order of their average $\kappa$ *over all four models* and then scored based on their rank in the sorted list. Only the top 5 methods received a score. For example, in Table 4 this scoring is applied to the aggregation methods for the algorithm classification dataset: *maxMed* (maximum concatenated with median) has the highest average over each model type and so gets a score of 5, *minMeanMax* gets score 4 and so on. This is repeated for all seven datasets thus giving a list of 35 pairs of aggregation functions and corresponding scores, with higher scores indicating the function performed better. These are then summed up for each aggregation method to establish their final score.

The final top 5 scoring aggregation functions are shown in Table 5. *mean* and *meanMin* obtained the highest score of 12, with *meanStd* performing almost as well with a score of 11.

Table 5: Top scoring aggregation functions over all 4 models.

| Aggregation Method | Sum of Score |
|---|---|
| mean | 12 |
| meanMin | 12 |
| meanStd | 11 |
| min | 9 |
| minMeanMax | 8 |

### 6.5 Obfuscation Comparison

We now consider the main research question: *Does obfuscation of variable names yield an improved model of code semantics?* To this end, we compare the embeddings created by the *Reduced* non-obfuscated model to those of the two models based on obfuscated variable names. Table 6 shows the $\kappa$ for each model on each dataset. The row shown for each dataset is based on the aggregation method that obtained the highest average $\kappa$ over all models. Green and red cells represent a statistically significant increase or decrease in performance, respectively, over the baseline (no obfuscation). All significance claims are based on $p < 0.05$ calculated by paired two-tailed t-test based on the estimates obtained from the corresponding 2×10 runs of stratified 10-fold cross-validation.

**OpenCV/Spring:** The OpenCV/Spring binary classification dataset was the easiest classification problem, showing the highest $\kappa$ overall of any dataset used in this experiment. The strong performance on this dataset indicates that the embeddings provided by the **code2vec** model cleanly separates these two different types of Java programs. The maximum $\kappa$ was 0.991 achieved by the *Random* model while the non-obfuscated model performed the worst with a $\kappa$ of 0.952. Comparing the average $\kappa$ between the baseline and *Random* model, the latter obtained a mean $\kappa$ coefficient 0.039 higher, which is a statistically significant difference ($\alpha = 9.87 \times 10^{-11}$).

**Algorithm Classification:** In the algorithm classification domain, the embeddings from the obfuscated model with *Random* obfuscation also performed the best. It obtained a .059 improvement in $\kappa$ over the baseline, which again is a statistically significant improvement ($\alpha = 4.50 \times 10^{-7}$). The model based on *Type* obfuscation also performed significantly better than the baseline with a 0.052 improvement, and an $\alpha$ of $7.24 \times 10^{-4}$.

This is a much harder dataset than OpenCV/Spring, but it is an appropriate task for the embeddings—the embeddings should reflect the semantics of code, so using them, for example, to classify between sorting and searching algorithms is appropriate—and the strong performance confirms this.

**Duplicate Detection:** Antithetically, the duplicate detection datasets saw less improvement provided by obfuscation. For the *duplicate file* detection task, the baseline performed marginally better on average with a $\kappa$ of 0.938. The *Type* obfuscated model showed a minor decrease in performance at 0.936, while the *Random* model decreased further at a $\kappa$ of 0.926. Despite this finding, for *duplicate function* detection the *Random* model performed statistically significantly *better* than the baseline with a $\kappa$ of 0.657, but the extent of the difference is relatively small (0.012). Overall, the embeddings were moderately suitable for this task: for detecting duplicate files, the embeddings performed fairly well.

Table 6: Average kappa coefficient over 10 runs of 10-fold CV, showing results for the most accurate aggregation method over all model types. Green/red cells indicate statistically significant increase/decrease over no obfuscation, respectively.

| Dataset | Best Aggregation | None | Type Obfuscation | Random Obfuscation |
| --- | --- | --- | --- | --- |
| OpenCV/Spring | minStd | 0.952 | 0.959 | **0.991** |
| Algorithm Classification | maxMed | 0.699 | 0.751 | 0.758 |
| Duplicate Files | medMin | **0.938** | 0.936 | 0.926 |
| Duplicate Functions | minMeanMax | 0.645 | 0.634 | **0.657** |
| Bug Detection | min | 0.265 | 0.272 | **0.285** |
| Malware Classification | mean | 0.448 | 0.424 | **0.449** |
| Author Attribution | meanStd | **0.261** | 0.223 | 0.200 |

**Bug Detection:** The results on the bug detection dataset were poor overall. Although each model performed poorly on this task in absolute terms, the *Random* model still produced a statistically significant improvement ($\alpha = 5.1 \times 10^{-4}$) over the baseline, with a mean $\kappa$ of 0.285 compared to 0.265. *Type*-based obfuscation gave a mean $\kappa$ in between (0.272).

The bug detection dataset is difficult for the embeddings because semantically similar code can appear in both classes (e.g., a sort implementation with a bug and without). As the embeddings were not fine-tuned to account for this, there is a large number of mispredictions. A more effective approach in this domain would be to fine-tune a pretrained **code2vec** model on this dataset. In theory, the attention mechanism in the neural network model would be able to focus on parts of code that are more bug-prone; the attention mechanism in our models is trained to focus on parts of the code that hold the semantics of the entire method, which does not necessarily correlate with whether the code has a bug or not.

**Malware Classification:** The performance between models on the malware classification dataset was fairly even. The embeddings trained with no obfuscation and random obfuscation performed in effect equally with $\kappa$ 0.448 and 0.449 respectively. The lack of improvement achieved by obfuscating could be because of the nature of malware code: malware creators often obfuscate their code preemptively to avoid detection. Additionally, the decompiler used to obtain the Java source code for the malware programs did not recover all variable names so the dataset was already semi-obfuscated.

The malware classification dataset is moderately suitable for the learned embeddings. Again, a more accurate approach would be to fine tune these embeddings for the malware classification domain. The learned embeddings are somewhat effective at classifying between different types of malware, but there is likely a lot of semantically common code between classes (e.g., most malware steal a device's information), which is one explanation for the lower accuracy compared to that observed in other domains.

**Author Attribution:** As expected, the embeddings for this dataset performed the worst overall. The non-obfuscated embedding model's mean $\kappa$ of 0.261 was significantly better than both the type and random obfuscated models, which obtained mean $\kappa$ of 0.223 ($\alpha = 3.18 \times 10^{-10}$) and 0.2 ($\alpha = 2.66 \times 10^{-21}$) respectively.

Understandably, variable names are useful for predicting code authorship: different programmers may use different naming conventions, so hiding this information makes the task significantly more difficult. The bias-reducing effect of hiding variable names that showed improvement in other domains fomented a detrimental result for the code author attribution dataset.

**Discussion:** The results on the seven datasets show obfuscation to be a useful strategy in regularising **code2vec**. Of the two types of obfuscation, random obfuscation is superior; additional significance testing shows that, over all seven datasets, type-obfuscated embeddings never provide a statistically significant improvement over random-obfuscated embeddings, while there are three datasets in which random-obfuscated embeddings are more predictive than type-obfuscated ones. Thus, the full benefit of obfuscation is obtained from complete variable obfuscation, where the **code2vec** model cannot learn any semantics based on variable names.

Embeddings trained on data with *Random* obfuscation improve on those trained without obfuscation, with statistically significant improvements in almost all of the appropriate tasks—tasks in which the functionality of the code is important for classification: OpenCV/Spring, Algorithm Classification, and Duplicate Detection. This indicates that training a **code2vec** model on randomly obfuscated data results in embeddings that generalize better. As shown in Section 6.1, obfuscating variable names **does** negatively affect the performance on the training task (method name prediction). Despite this, the quality of the embeddings provided by these obfuscated models is generally **not** negatively affected, and in some cases usefully improved.

These results indicate that code2vec *code embeddings* are most representative when ignoring *token embeddings* for variable names, even if these are generic names as obtained with type obfuscation.

## 6.6 Training with Complete Information

As mentioned earlier, the developers of **code2vec** investigated hiding information *at test time* by masking parts of the path-context. They trained a model with complete information and tested it with partial information, which unsurprisingly gave poor results.

Conversely, this project investigates the effect of training *and* testing with hidden information. To illustrate the difference and how hiding information at test time severely impairs a **code2vec** model, a set of experiments were performed in which a non-obfuscated **code2vec** model was trained and then evaluated in non-obfuscated and obfuscated contexts for method name prediction. As we are using the task of method name prediction, we report the F1 score returned by the code2vec model during evaluation.

Results are shown in Table 7, which demonstrate that the two non-obfuscated models (Standard and Reduced) suffer a strong decline in prediction performance when obfuscating the method-name prediction test data (using random obfuscation), dropping below that of the *Random*; in contrast, the *Random* model exhibited no significant difference in performance whether variable names were visible or not. This drop in performance provides additional evidence that the non-obfuscated models rely heavily on variable names: the non-obfuscated models perform worse than the *Random* model on the obfuscated data, as a source of information they learned to leverage (variable names) is hidden. Hiding variable names made little difference to the *Random* model, as the attention mechanism was trained to focus on other parts of the source code. As mentioned in Section 2, embeddings from a non-obfuscated model can change drastically based only a subtle change in variable naming; the *Random* model does not have this issue because variable names are obfuscated anyway.

**Table 7: F1 score for the task of method prediction, obtained on obfuscated and non-obfuscated testing data.**

| Test Dataset Type | Standard | Reduced | Random |
|---|---|---|---|
| No Obfuscation | **0.433** | 0.431 | 0.384 |
| Random Obfuscation | 0.343 | 0.343 | **0.386** |

## 7 CONCLUSIONS AND FUTURE WORK

Applying machine learning to static code analysis is an exciting area of research, allowing the large amount of free source code available today to help solve tasks like plagiarism and malware detection. However, as further methods are researched for extracting information from large corpora of source code in the future it is imperative that they can generalise effectively and are robust at inference-time to surface-level changes in the code. Training on any large-scale crowd-sourced data should be approached carefully as biases in the data can be amplified in the models, causing them to be fooled easily and generalise poorly; this was shown to be the case for **code2vec**. To address this, we investigated approaches for reducing reliance on variable names by training a **code2vec** model on obfuscated data, which forces the model to look at the other structure in the code. We have also expanded the application of the model to entire classes of Java code: we presented simple mathematical operations for aggregating sets of vectors into a single vector and compared them empirically.

Our results show that obfuscation of variable names reduces **code2vec** performance for the method-name prediction task that is used for training the model; however, the embedding evaluation shows that it better preserves code semantics.

Regarding selection methods for deriving class-level embeddings through aggregation, we found that it is best to simply use all method embeddings. Similarly, dimensionality reduction of the embeddings using UMAP did not prove useful. Regarding the method used for aggregating per-method embeddings, we have found that there is no single best aggregation method: a combination of *min/max* and *mean/median/std dev* keeps the most information about the set of vectors.

For obfuscating variable names, the random method provided a larger number of significant improvements than type-based obfuscation, indicating that completely removing reliance on variable names is preferable as it completely removes the potential of the model to overfit to variable names, forcing it to rely on method names, literals and the code's structure. In most cases this will provide an improvement, or at least is unlikely to cause a significant deterioration. Consequently, for the general case, **code2vec** models should be trained on obfuscated code. However, for tasks where *variable names* are crucial, such as author attribution, **code2vec** should be trained *with* variable names (**code2vec**'s current training task is unsuitable for code author attribution, as explained in Section 5.3).

This research has shown variable obfuscation as a potential regularization process for machine learning on source code, making the resulting model more robust with less bias towards variable names, culture agnostic—not affected by how variables are named or the language the variables are named in—and importantly, more effective in tasks where variable names may not be reliable at all.

The datasets we collated for this research are a valuable resource for testing code embeddings as they cover a wide range of tasks and highlight the limitations of the **code2vec** embeddings (e.g., author attribution). All of these are made available (with the exception of the malware classification dataset, due to ethical concerns) for further research and benchmarking of machine learning approaches to static code analysis tasks, and can be found at this paper's file repository[10].

The results show that **code2vec** does not provide an optimal representation for every possible task. Method-name prediction is a novel use of the large amount of source code available and is the current state-of-the-art for creating general purpose embeddings, but it is not clear whether different training objectives (e.g. class name prediction) would yield more effective representations. Method-name prediction may also not be a suitable training task over different representations (e.g. graph-based models).

Additionally, There are several other promising avenues for future work. For example, we have only considered basic methods for aggregating per-method embeddings. A more sophisticated approach would be to use a neural attention approach to combine the individual method embeddings, such as the one applied by **code2vec** itself to learn a weighting when combining context vectors.

Additionally, a semi-obfuscated **code2vec** model could be trained, where only 50% of variables are obfuscated or the original dataset is augmented with obfuscated versions of the same code. This may create a model that can use variable names when they are informative, but does not rely heavily on doing so and can still generalise well. It is also important to note that we have not considered obfuscating method names alongside variable names; this is an interesting avenue for future work.

Finally, to provide more information to the model, aggressive inlining of method calls could be applied to the code. This removes the abstraction and allows the model to see exactly what the invoked code is doing rather than just the—potentially undescriptive—method name.

---

[10]https://github.com/basedrhys/obfuscated-code2vec